\documentclass[runningheads]{llncs}
\usepackage{graphicx}
\usepackage[utf8]{inputenc} 
\usepackage[T1]{fontenc}    
\usepackage{hyperref}       
\usepackage{url}            
\usepackage{booktabs}       
\usepackage{amsfonts}       
\usepackage{nicefrac}       
\usepackage{microtype}      
\usepackage{xcolor}         
\usepackage{graphicx}
\usepackage{algorithm}
\usepackage{algpseudocode}
\usepackage{amsmath}
\usepackage{amssymb}
\usepackage{comment}
\usepackage{multirow}
\usepackage{caption}
\usepackage{hyperref}

\usepackage{marvosym}
\usepackage{ifsym}

\makeatletter
\newcommand*{\myfnsymbolsingle}[1]{%
  \ensuremath{%
    \ifcase#1
    \or 
      \dagger
    \or 
      \dagger
    \or 
      \dagger
    \or 
      \dagger
    \or 
      \dagger
    \or 
      \dagger
    \or 
      \dagger
    \or 
      \dagger
    \else 
      \@ctrerr  
    \fi
  }%
}   
\makeatother

\usepackage{alphalph}
\newalphalph{\myfnsymbolmult}[mult]{\myfnsymbolsingle}{}

\usepackage{fnpct} 

\begin{document}
\title{LarvSeg: Exploring Image Classification Data For Large Vocabulary Semantic Segmentation via Category-wise Attentive Classifier\thanks{Supported by National Key R \& D Program of China (2022ZD0160300) and National Science Foundation of China (NSFC62376007).}}
\titlerunning{LarvSeg: Category-wise Attentive Classifier for Image Segmentation}
\author{Haojun Yu\inst{1,\dagger}\orcidID{0009-0004-5291-8363} \and
Di Dai\inst{1,\dagger}\orcidID{0009-0002-1832-5164} \and
Ziwei Zhao\inst{2}\orcidID{0000-0003-1160-5737} \and
Di He\inst{1(\textrm{\Letter})}\orcidID{0009-0003-0468-1288} \and
Han Hu\inst{3}\orcidID{0000-0001-5104-6146} \and
Liwei Wang\inst{1,2}\orcidID{0000-0003-1739-2621}
}

\authorrunning{H. Yu et al.}

\institute{National Key Laboratory of General Artificial Intelligence, School of Intelligence Science and Technology, Peking University, Beijing, China\\
\email{dihe@pku.edu.cn}
\and
Center of Data Science, Peking University, Beijing, China \and
Microsoft Research Asia, Beijing, China\\
}
\maketitle  
\begin{abstract}
Scaling up the vocabulary of semantic segmentation models is extremely challenging because annotating large-scale mask labels is labour-intensive and time-consuming. Recently, language-guided segmentation models have been proposed to address this challenge. However, their performance drops significantly when applied to out-of-distribution categories. In this paper, we propose a new large vocabulary semantic segmentation framework, called LarvSeg. Different from previous works, LarvSeg leverages image classification data to scale the vocabulary of semantic segmentation models as large-vocabulary classification datasets usually contain balanced categories and are much easier to obtain. However, for classification tasks, the category is image-level, while for segmentation we need to predict the label at pixel level. To address this issue, we first propose a general baseline framework to incorporate image-level supervision into the training process of a pixel-level segmentation model, making the trained network perform semantic segmentation on newly introduced categories in the classification data. We then observe that a model trained on segmentation data can group pixel features of categories beyond the training vocabulary. Inspired by this finding, we design a category-wise attentive classifier to apply supervision to the precise regions of corresponding categories to improve the model performance. Extensive experiments demonstrate that LarvSeg significantly improves the large vocabulary semantic segmentation performance, especially in the categories without mask labels. For the first time, we provide a 21K-category semantic segmentation model with the help of ImageNet21K. The code is available at \href{https://github.com/HaojunYu1998/LarvSeg}{https://github.com/HaojunYu1998/LarvSeg}.

\end{abstract}

\section{Introduction}
\label{sec:introduction}

\stepcounter{footnote}\footnotetext{Equal contribution.}

\begin{figure}[t]
    \centering
    \includegraphics[width=\linewidth]{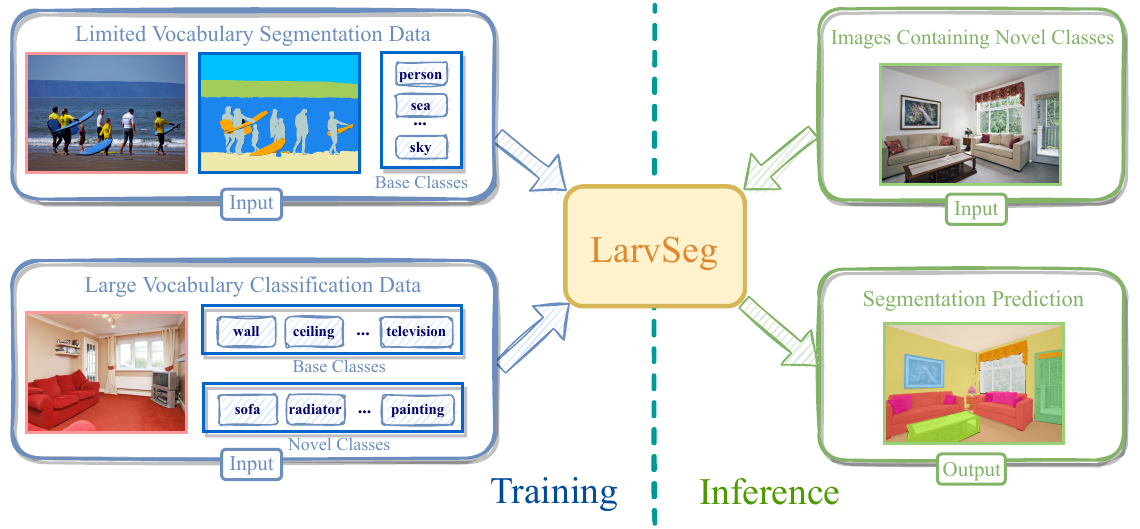}
    \captionsetup{justification=raggedright,singlelinecheck=false}
    \caption{Illustration of a new paradigm to address large vocabulary semantic segmentation with image classification data.}
    \label{fig:teaser}
\end{figure}

Semantic segmentation is a fundamental visual task that aims to assign a semantic class to each pixel in the image. Recently, deep learning-based methods have achieved great success with a supervised learning paradigm for semantic segmentation~\cite{open2023,learn2023,long2015fully,strudel2021segmenter,chen2017deeplab,cheng2021per,badrinarayanan2017segnet,yuan2020object}. In real applications, the segmentation tasks often focus on a large set of given categories (large vocabulary). For example, the segmentation of various vehicles is a necessary module for autonomous driving. However, obtaining pixel-level mask labels is labour-intensive and time-consuming: annotating one accurate mask of an object typically requires 54 to 79 seconds for a well-skilled annotator~\cite{bearman2016s}. Thus the vocabulary of widely used semantic segmentation datasets~\cite{zhou2017scene,caesar2018coco,zhou2017scene,cheng2021per} is limited, only comprising a few hundreds of categories. Segmentation models trained on these datasets fail to recognize the given categories that are out of the training distribution (OOD).

Recently, a new paradigm called language-guided segmentation~\cite{zhou2021denseclip,li2022language,xu2021simple,ghiasi2021open,Kirillov_2023_ICCV} has been proposed to address this challenge. These works aim to extend the segmentation vocabulary by leveraging language semantics. For example, models trained on "cat" can correctly segment the unseen category "furry" because they are semantically similar in language space. Ideally, this paradigm can perform segmentation on any category (open vocabulary). However, the model performance significantly drops when facing OOD text prompts(see Table~\ref{tab:sota_comp}). The key reason is that \textbf{vision encoders has never received the supervision of semantics similar to OOD text prompts}, thus visual features are not aligned well with the OOD text features in the language space.

In this paper, we address the challenge of large vocabulary semantic segmentation with a novel framework, LarvSeg. The key idea is to introduce image classification data as coarse supervision of concerned semantics to significantly improve model performance. For clear discussion, we denote categories with pixel-level or image-level labels as \texttt{base}, \texttt{novel} categories respectively. Initially, we developed a training strategy to train a simple baseline with both image-level and pixel-level supervision. Surprisingly, we find that \textbf{ pixels of \texttt{novel} categories in the feature maps have already been clustered when training the segmentation model with only \texttt{base} categories}. Based on this key observation, we propose a category-wise attentive classifier (CA-Classifier) to apply category supervision to the precise regions.

We extensively evaluate our method on a wide range of tasks.  We use COCO-Stuff~\cite{caesar2018coco} as the limited vocabulary segmentation dataset, use ImageNet21K~\cite{deng2009imagenet} (21,841 categories), ADEFull~\cite{zhou2017scene,cheng2021per} (847 categories) and ADE20K~\cite{zhou2017scene} (150 categories) as image-level classification datasets in the experiments, where the statistics of all datasets are summarized in Table~\ref{tab:datasets}. We jointly train segmentation models with C171 and different image classification datasets: 
(1) multi-label classification datasets like WA847; 
(2) combination of single- and multi-label classification datasets like I585 and WA847;
(3) ImageNet21K~\cite{deng2009imagenet} for 21K-category semantic segmentation.
The experimental results show that the proposed simple baseline surpasses previous open vocabulary arts~\cite{zhou2021denseclip,li2022language,xu2021simple,ghiasi2021open} by a large margin, and LarvSeg outperforms the baseline on \texttt{novel} categories by $6.0$ mIoU on A150 and $2.1$ mIoU on A847.

Our contributions to tackle the challenge of large vocabulary semantic segmentation are four-fold. (1) We identify the key weakness of language-driven segmentation models is the lack of OOD supervision. (2) We propose a new framework leveraging classification data as coarse semantic supervision. (3) We propose a category-wise attentive classifier to further improve the model performance, especially on \texttt{novel} categories. (4) For the first time, we provide a model to perform semantic segmentation on 21K categories with the help of ImageNet21K.

\section{Related Work}
\label{sec:related_works}

\noindent\textbf{Large Vocabulary Object Detection} This task utilizes image classification data to extend object detection to a large vocabulary. Detic~\cite{zhou2022detecting} trains the classification branch of a detector with ImageNet21K~\cite{deng2009imagenet} utilizing the RoI features of the largest proposal. Simple-21K-Detection~\cite{lin2022simple} proposes to transfer knowledge from an image classifier to object detection data and adopt a two-stage approach to achieve large vocabulary detection. Both of them can detect objects in 21K categories. Our work shares the same motivation with these works and for the first time extends the vocabulary of the semantic segmentation task to 21K categories.

\noindent\textbf{Language Guided Semantic Segmentation} This task aims to perform semantic segmentation on images with arbitrary categories. Vision language models like CLIP~\cite{radford2021learning} or ALIGN~\cite{jia2021scaling} trained with web-scale image-text pairs~\cite{schuhmann2021laion,thomee2016yfcc100m,changpinyo2021cc12m} provide a coarse alignment between image and text. Previous works~\cite{li2022language,xu2021simple,ghiasi2021open,shin2022reco,xu2022groupvit,ding2022open,Kirillov_2023_ICCV} take advantage of the coarse alignment to endow segmenters with the ability to perform semantic segmentation on arbitrary categories. However, as the models have not received any supervision of unseen categories, their performance in unseen categories is much worse than in seen categories. Instead of pursuing open vocabulary semantic segmentation, in this paper, we perform segmentation on a large (but closed) vocabulary provided by classification data. Therefore, our framework to bridge the gap between image-level and pixel-level labels is complementary to language-guided methods. 

\section{Method}
\label{sec:method}

\subsection{A Simple Baseline}
\label{subsec:baseline}

\noindent\textbf{Incorporating Image-Level Supervision}\label{def:img_level_score}
Our basic segmenter contains a backbone network (e.g. ViT-B/16~\cite{dosovitskiy2020image}) followed by a one-layer cosine classifier to predict pixel-wise categories. To incorporate image-level supervision into the training process, we conduct an image-level classification of the average pooled feature map. For multi-class images, the cross entropy loss is calculated separately for each category in the image. To avoid conflicts among different categories, we set other categories in the image as ignore labels, which indicates these categories are not involved in the actual calculation. 

\noindent\textbf{Joint Training} To enable the segmentation model to recognize \texttt{novel} categories, we jointly train the network with segmentation and classification data. The model parameters are shared for both tasks. The overall training objective is: 
\begin{align}
\mathcal{L}=\mathcal{L}_{\text{seg}}+\lambda_{\text{cls}} \mathcal{L}_{\text{cls}},
\end{align}
where the loss weight $\lambda_{\text{cls}}$ is a hyperparameter which is set as $0.1$ by default.

\noindent\textbf{Inference} During inference, the predicted categories for each pixel are directly obtained as the original segmentation model. Since the model parameters are shared for both tasks, the segmenter naturally possesses the ability to segment \texttt{novel} categories that appeared in image classification data.

\subsection{Analysis For Novel Category Pixel Grouping}
\label{subsec:subtask_analysis}

\begin{figure}[h]
    \centering
    \includegraphics[width=0.7\linewidth]{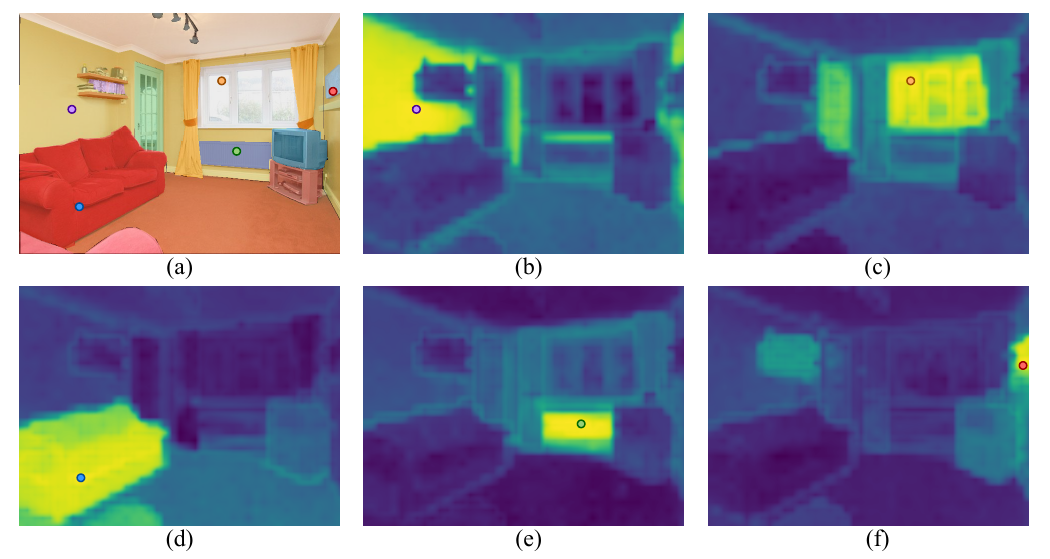}
    \caption{Visualization of the response maps for pixel grouping. The model is trained on C171 and the response maps are visualized on A150. The dots with different colours are the selected pixels for different categories. (a) denotes the image overlayed with the ground truth mask; (b) and (c) is the response maps of \emph{wall} and \emph{window}, which are inside the training vocabulary; (d), (e) and (f) are the response maps of \emph{sofa}, \emph{radiator} and \emph{painting}, which are outside training vocabulary. We observe that (d), (e) and (f) present intra-category compactness as good as (b) and (c).}
    \label{fig:pixel_group}
\end{figure}

\begin{table}[b]
    \setlength{\tabcolsep}{3pt}
    \small
    \centering
    \begin{tabular}{l|lc|lc}
        \toprule
        Model & Training Data & A150$-$C171 & Training Data & A847$-$C171 \\
        \midrule
        \midrule
        DenseCLIP & ITPair & 21.6 & ITPair & 11.2\\ 
        Basic Segmenter & C171 & 48.6 & C171 & 36.1\\
        Basic Segmenter & A150 & \color{gray}{49.0} & A847 & \color{gray}{37.8}\\
        \bottomrule
    \end{tabular}
    \caption{Results of the exploratory experiment. For the second row, the evaluation categories are not in the training vocabulary. For the third row, the evaluation categories are within the training vocabulary. CLIP-RN50x64 with VILD prompt engineering is used to generate DenseCLIP~\cite{zhou2021denseclip} results.}
    \label{tab:oracle_exp}
\end{table}

Despite the good performance, the proposed baseline is faced with potential limitations. The probability vector $\bar{\mathcal{P}}_{\text{cls}}$ of the image is obtained through global average pooling, indicating that the classification supervision has not been applied to the corresponding pixels accurately. To alleviate this problem, we intend to extract foreground regions of \texttt{novel} categories and apply fine-grained supervision to them. For precise region extraction without pixel-level mask labels, intra-category pixel compactness is essential. Thus, we train a segmentation model with \texttt{base} categories and investigate the pixel feature similarity of \texttt{novel} categories.

\noindent\textbf{Exploratory Experiment}  This experiment is designed to measure the intra-category pixel compactness in a category-agnostic way. Specifically, we train a basic segmenter with only \texttt{base} categories. During inference, for images containing \texttt{novel} category regions, we extract feature maps $\mathcal{F}$ and randomly select a pixel $(h_c,w_c)$ from the ground truth mask of each category. The feature of the selected pixel is regarded as the representative feature of the category. Then, we calculate the cosine similarity between the representative feature $\mathcal{F}[h_c,w_c]$ and other pixels in feature maps to obtain the pixel-level response maps $\mathcal{S}_{\text{res}}$: 
\begin{align}
\label{eq:response_map}
    \mathcal{S}_{\text{res}}=\operatorname{cos}(\mathcal{F}[h_c,w_c], \mathcal{F}).
\end{align}
Higher scores in the response map indicate higher similarities with the selected representative feature of the category. Thus, the predicted category of each pixel is the category with the highest score. For an intuitive understanding, we visualize the response maps of different categories in Figure~\ref{fig:pixel_group}. The dots with different colors are the selected pixels for different categories and the response maps indicate the pixel similarities. The experimental results are shown in Table~\ref{tab:oracle_exp} which indicate that the intra-category pixel compactness of the categories outside the training vocabulary is almost the same as the categories inside the training vocabulary.

\begin{figure}[t]
    \centering
    \includegraphics[width=\linewidth]{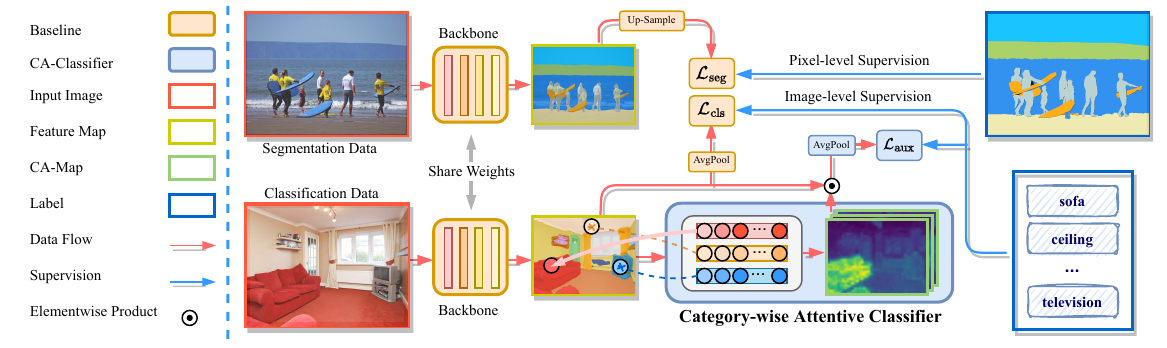}
    \caption{Illustration of LarvSeg framework. The meaning of each icon is listed on the left. CA-Classifier and CA-Map stand for category-wise attentive classifier and category-wise attention map defined in Section~\ref{def:ca_map}. The proposed simple baseline learns from segmentation and image classification data simultaneously via pixel-level and image-level classification tasks (the losses are denoted as $\mathcal{L}_{\text{seg}}$ and $\mathcal{L}_{\text{cls}}$ in the figure). Additionally, the proposed category-wise attentive classifier maintains category-wise features with a memory bank to highlight the foreground pixel group and suppress background pixel groups. The attentively pooled score map is supervised by an auxiliary image-level classification task (the loss is denoted as $\mathcal{L}_{\text{aux}}$ in the figure).}
    \label{fig:framework}
\end{figure}

\noindent\textbf{Discussion} We observe the pixel grouping ability is generalizable to out-of-distribution categories in a category-agnostic way. This is probably because pixels with the same semantic category are often similar in locations, colours, textures, etc. This experiment implies that \textbf{the pixel features of image classification data have already been clustered into implicit groups}. Fine-grained supervision signals can be generated from image-level labels if pixel groups of specific categories can be distinguished from other groups explicitly. 

\subsection{Category-wise Attentive Classifier}\label{subsec:background} 
Based on the above observations, we propose a Category-wise Attentive Classifier (CA-Classifier) to explicitly distinguish different pixel groups and extract the corresponding regions of \texttt{novel} categories to apply fine-grained supervision. We will elaborate on the details in the following.

\noindent\textbf{Memory Based Category Representation} 
To explicitly distinguish the pixel groups in a category-aware way, we employ cross-image semantic cues. During the training process, we maintain a memory bank to store representative features of each \texttt{novel} category $
\mathbb{M}=\{\mathcal{M}_{i}|\forall i\in \text{\texttt{novel} category}\}$. In each training step, we update the memory bank using pixel features with the top $K$ classification scores. The confidence score $\mathcal{S}^{i}_{\text{mem}}$ of each pixel belonging to each category is defined as the average cosine similarity between the pixel feature $\mathcal{F}_{(h,w)}$ and the representative features of the category in the memory bank:
\begin{align}
    \mathcal{S}^{i}_{\text{mem}} = \operatorname{Avg}\left(\cos(\mathcal{F}_{(h,w)}, \mathcal{M}_{i})\right),\forall(h,w).
\end{align}

\noindent\textbf{Category-wise Attention Map}\label{def:ca_map} 
The category-wise attention map is generated by foreground strengthening and background suppression. For the foreground category, we define a category-wise attention map $\mathcal{A}$ as the rescaled confidence score map using a sigmoid function:
\begin{align}
    \mathcal{A} = \operatorname{sigmoid}(\operatorname{norm}(\mathcal{S}_{\text{mem}}^{\text{fg}} - \mathcal{S}_{\text{mem}}^{\text{bg}}))
\end{align}
where $\text{fg}$ and $\text{bg}$ represent foreground and background category in the given image. Then, we use the category-wise attention map to attentively pool the feature map and apply an auxiliary image-level classification task to the attentively pooled feature map as in Section~\ref{subsec:baseline}:
\begin{align}\label{eq:odot}
    \mathcal{P}^{i}=\operatorname{softmax}(\mathcal{S}^{i}_{\text{cls}}\cdot\mathcal{A}/\tau), \forall i\in\text{\texttt{novel} category}
\end{align}
where $\mathcal{S}_{\text{cls}}$ is original classification score map. More details can be found in Supplementary Material. In this way, the foreground regions in the category-wise attention map will receive more attention as shown in Figure~\ref{fig:framework}, while the attention scores in background regions are close to zero. Thus, the image-level supervision will be accurately applied to the foreground pixel group. 

\noindent\textbf{Learning Objective} 
The overall loss function consists of the semantic segmentation loss $\mathcal{L}_{\text{seg}}$, the image classification loss $\mathcal{L}_{\text{cls}}$ and the auxiliary classification loss $\mathcal{L}_{\text{aux}}$ with category-wise attentive classifier:
\begin{align}
\mathcal{L}=\mathcal{L}_{\text{seg}}+\lambda_{\text{cls}}\mathcal{L}_{\text{cls}}+\lambda_{\text{aux}}\mathcal{L}_{\text{aux}},
\end{align}
where the loss weights $\lambda_{\text{cls}}$ and $\lambda_{\text{aux}}$ are hyperparameters which are both set as $0.1$ by default. The overall framework is presented in Figure~\ref{fig:framework}.

\subsection{Segmentation For 21K Classes}
\label{subsec:21k_seg}

To further tap the potential of the proposed method, we show that it can extend the vocabulary to 21K categories. Specifically, we employ ImageNet21K~\cite{deng2009imagenet} as the classification dataset to train LarvSeg. The qualitative results are presented in Section~\ref{subsec:qualitative_results}. More details and results can be found in Supplementary Material.

\section{Experiments}
\label{sec:experiments}

\subsection{Datasets}
\label{subsec:datasets}
In this subsection, we will introduce the datasets used in this paper. The summary of all datasets is shown in Table~\ref{tab:datasets}.

\noindent\textbf{C171.} COCO-Stuff~\cite{caesar2018coco} denoted as C171 includes 118k/5k train/val images and 171 categories with pixel-level labels. We employ C171 as training data to provide pixel-level supervision. Other classification datasets with image-level labels are used to extend the C171 vocabulary.

\noindent\textbf{A150 and WA150.} ADE20K~\cite{zhou2017scene} denoted as A150 includes 20k/2k train/val images and 150 categories with pixel-level labels. We use the A150 validation set as an evaluation benchmark. We denote A150 with only image-level labels as WA150 (weakly supervised A150). The WA150 training set is used as a multi-label classification dataset to extend the C171 vocabulary. WA150 contains 87 \texttt{novel} categories that are not included in C171.

\noindent\textbf{A847 and WA847.} ADEFull~\cite{zhou2017scene,cheng2021per} denoted as A847 includes 20k/2k train/val images and 847 categories. We use the A847 validation set as an evaluation benchmark. We denote A847 with only image-level labels as WA847 (weakly supervised A847). WA847 training set is used as a multi-label classification dataset to extend the C171 vocabulary. WA847 contains 755 \texttt{novel} categories that are not included in C171.

\noindent\textbf{I21K, I124 and I585.} ImageNet21K~\cite{deng2009imagenet} denoted as I21K includes 14M images and 21K categories. It is a single-label classification dataset, and the objects are usually large and located in the image centre. The intersection of the 21K categories with A150 categories contains 124 categories and we collect the I21K images with these 124 categories to construct I124. Similarly, we construct I585 with the images belonging to the 585 intersected categories between I21K and A847. 

\begin{table}[t]
    \centering
    \small
    \setlength{\tabcolsep}{3pt}
    \begin{tabular}{llll}
    \toprule
    Notation & Definition & \#Img & \#Cls\\
    \midrule
    \midrule
        C171 & COCOStuff dataset & 118K+5K & 171\\
        A150  & ADE20K dataset & 20K+2K & 150\\
        A847  & ADEFull dataset & 20K+2K & 847\\
        A150-C171 & A150 validation set with A150$-$C171 categories & 2K & 87\\
        A847-C171 & A847 validation set with A847$-$C171 categories & 2K & 809\\
        WA150  & ADE20K training set with classification labels & 20K & 150\\
        WA847  & ADEFull training set with classification labels & 20K & 847\\
        I21K  & ImageNet21K dataset & 14M & 21K\\
        I124    & I21K with I21K$\cap$A150 categories & 174K & 124\\
        I585    & I21K with I21K$\cap$A847 categories & 739K & 585\\
        LN & Localized Narratives dataset & 652K & - \\ 
        ITPair  & Web-scale Image-Text pairs & 400M-800M & - \\
    \bottomrule
    \end{tabular}
    \captionsetup{singlelinecheck=off}
    \captionsetup{justification=raggedright}
    \caption{Abbreviations, brief introduction and vocabulary sizes of the datasets used in this paper.}
    \label{tab:datasets}
\end{table}

\subsection{Experimental Setups}
\label{subsec:exp_settings}

\noindent\textbf{Weak-ADE As Classification Data} We extend the vocabulary of C171 with the multi-label classification dataset WA150 (WA847) and evaluate the performance on A150 (A847). The multi-label classification datasets are originally formed as segmentation data, so they are imbalanced in categories.

\noindent\textbf{ImageNet and Weak-ADE As Classification Data} We further add I124 (I585) to the above settings and evaluate the models on A150 (A847). The single-label ImageNet dataset is adequate and category-balanced compared to WA150 and WA847.

\noindent\textbf{21K-category Segmentation} We extend the vocabulary of semantic segmentation models to 21K categories with dataset I21K. In Supplementary Material, we provide qualitative results to demonstrate the model's ability to segment \texttt{novel} categories.

\subsection{Implementation details}
\label{subsec:impl_details}

\noindent\textbf{Overall Framework} We use ViT-B/16~\cite{dosovitskiy2020image} initialized with I21K pre-trained weights as the backbone by default. All the models are trained on $8$ V100 GPUs. Following Detic~\cite{zhou2022detecting}, we group images from the same dataset on the same GPU to improve training efficiency. For all experiments, we sample two images in a mini-batch for each GPU. For setups (1) and (3), we use $4: 4$ GPUs for segmentation and classification data. For setup (2), we use $3: 3: 2$ GPUs for C171, WA847 (WA150) and I585 (I124).

\noindent\textbf{Training Parameters} Models for the main results and ablation study are trained for $320$K iterations. For the 21K categories segmentation, we train the model for 1280K iterations. Images from C171, WA150 and WA847 are resized to $512\times 512$ and images from I21K, I124 and I585 are resized to $320\times 320$. We apply multi-scale jittering with a random scale between $[0.5, 2.0]$, random crop, random flip and photometric distortion as data augmentations. SGD is adopted as the optimizer with a base learning rate of $0.001$, momentum $0.9$ and no weight decay. We use the polynomial learning rate schedule with power $0.9$ and minimum learning rate $1e^{-5}$. We empirically set the loss weights $\lambda_{\text{cls}}=0.1$ and $\lambda_{\text{aux}}=0.1$. All results are evaluated without multi-scale tests.

\subsection{Main Results}
\label{subsec:results}

\begin{table}[t]
\tiny 
\centering\setlength{\tabcolsep}{2pt}
    {
    \begin{tabular}{ll|lccc|lccc}
        \toprule
        \multirow{2}{*}{Model} & \multirow{2}{*}{Backbone} & \multirow{2}{*}{Training Data} & \multicolumn{3}{c|}{A150-mIoU} & \multirow{2}{*}{Training Data} & \multicolumn{3}{c}{A847-mIoU}\\
        & & & All & Base & Novel & & All & Base & Novel\\ 
        \midrule
        \midrule
        DenseCLIP~\cite{zhou2021denseclip} & RN50x64~\cite{he2016deep} & ITPair & 11.5 & 15.4 & 9.1 & ITPair & 4.5 & 8.8 & 3.9 \\
        ReCo~\cite{shin2022reco} & RN50x64 & I1K, ITPair & 12.6 & 15.8 & 10.5 & I1K, ITPair & 5.0 & 9.2 & 4.5 \\
        ZSBaseline~\cite{xu2021simple} & ViT-B/16~\cite{dosovitskiy2020image} & C171, ITPair & 17.7 & 31.0 & 9.3 & C171, ITPair & 5.2 & 14.7 & 4.1 \\
        LSeg~\cite{li2022language} & ViT-B/16 & C171 & 19.3 & 40.2 & 6.8 & C171 & 3.9 & 25.8 & 1.2 \\
        OpenSeg~\cite{ghiasi2021open} & EfficientNet-B7~\cite{tan2019efficientnet} & C171, LN, ITPair & 24.8 & - & - & C171, LN, ITPair & 6.8 & - & - \\
        \midrule
        Baseline & ViT-B/16 & C171, WA150 & 28.5 & 40.9 & 20.8 & C171, WA847 & 8.9 & 28.7 & 6.6\\
        LarvSeg & ViT-B/16 & C171, WA150 & \textbf{32.4} & \textbf{41.3} & \textbf{26.8} & C171, WA847 & \textbf{10.9} & \textbf{29.0} & \textbf{8.7}\\
        \midrule
        Baseline & ViT-B/16 & C171, WA150, I124 & 31.3 & 40.7 & 25.3 & C171, WA847, I585 & 11.9 & 28.5 & 9.9\\
        LarvSeg & ViT-B/16 & C171, WA150, I124 & \textbf{33.7} & \textbf{41.3} & \textbf{28.8} & C171, WA847, I585 & \textbf{13.2} & \textbf{30.4} & \textbf{11.1}\\
        \midrule
        Supervised & ViT-B/16 & A150 & \color{gray}{48.6} & \color{gray}{56.2} & \color{gray}{43.7} & A847 & \color{gray}{18.6} & \color{gray}{40.6} & \color{gray}{15.9}\\
        \bottomrule
    \end{tabular}
    }
    \caption{\textbf{Comparison of our proposed baseline and LarvSeg with open vocabulary state-of-the-art models and supervised learning models.} The supervised learning results are presented in \color{gray}{gray}\color{black}. ITPair stands for the web-scale image-text pairs to train a CLIP/ALIGN model. ReCo results are generated using the official checkpoint. ZSBaseline results are our reproduction with ViT-B/16 backbone for a fair comparison. LSeg is our reproduction with ViT-B/16 backbone and trained on C171 for a fair comparison. All models are evaluated without multi-scale tests and Dense CRF post-process.}
    \label{tab:sota_comp}
\end{table}

\noindent\textbf{Weak-ADE As Classification Data.} In Table~\ref{tab:sota_comp}, we compare our new paradigm with state-of-the-art results in language-guided segmentation paradigm~\cite{zhou2021denseclip,shin2022reco,li2022language,xu2021simple,ghiasi2021open}. Our simple baseline outperforms OpenSeg\cite{ghiasi2021open} by a large margin. Note that OpenSeg is initialized with ALIGN~\cite{jia2021scaling} weights (about 800M image-text pairs) and trained on C171 and Localized Narratives~\cite{PontTuset_eccv2020} (about 652K image-text pairs). What's more, LarvSeg can bring a significant improvement to the baseline model ($2.1$ mIoU in A847 \texttt{novel} categories; $6.0$ mIoU in A150 \texttt{novel} categories;). This result indicates the effectiveness of the proposed category-wise attentive classifier.

\noindent\textbf{ImageNet and Weak-ADE As Classification Data.} In Table~\ref{tab:sota_comp}, we additionally include single-label classification data I124 (I585) in the training set. The simple baseline significantly improves the performance ($3.3$ mIoU in A847 \texttt{novel} categories; $4.5$ mIoU in A150 \texttt{novel} categories). When adding the CA-Classifier, we observe a consistent $1.2$ mIoU improvement in A847 \texttt{novel} categories; and $3.5$ mIoU improvement in A150 \texttt{novel} categories.

\noindent\textbf{Discussion.} Utilizing ImageNet data, we consistently improve the model performance in all settings. The reasons are three-fold. (1) ImageNet is adequate and category-balanced compared to Weak-ADE, addressing the challenge of insufficient supervision of tail categories as shown in Supplementary Material. (2) Using additional datasets allows the model to learn from diverse visual appearances which improves the generalization ability. (3) The single-label images from ImageNet21K are object-centric with fewer background regions, which naturally provide more accurate supervision. This result indicates that besides model architecture design, \textbf{collecting additional high-quality data is the key way to improve the model performance} instead of designing model architectures.

\subsection{Ablation Study}
\label{subsec:ablation}

\begin{table}[t]
\small
\centering\setlength{\tabcolsep}{5pt}
    {
    \begin{tabular}{cccccc}
        \toprule
        CAC-WA150 & CAC-I124 & All & Base & Novel\\
        \midrule
        \midrule
        - & - & 31.3 & 40.7 & 25.3 \\
        - & $\checkmark$ & 32.1 & 40.8 & 26.6\\
        $\checkmark$ & - & 33.1 & 41.0 & 28.2\\
        $\checkmark$ & $\checkmark$ & \textbf{33.7} & \textbf{41.3} & \textbf{28.8}\\
        \bottomrule
    \end{tabular}
    }
    \caption{The A150-mIoU results of CA-Classifier for single- and multi-label supervision respectively. CAC-WA150 with $\checkmark$ means using CA-Classifier on WA150 data and others are similarly defined.}
    \label{tab:ca_cls_effective}
\end{table}

\begin{table}[t]
\small
\centering\setlength{\tabcolsep}{5pt}
    {
    \begin{tabular}{lccccc}
        \toprule
        CA-Classifier & All & Base & Novel\\
        \midrule
        \midrule
        Single-Image & 30.1 & 38.6 & 24.8 \\
        Cross-Image & \textbf{32.4} & \textbf{41.3} & \textbf{26.8}\\
        \bottomrule
    \end{tabular}
    }
    \caption{The A150-mIoU results of different designs for CA-Classifier. Single-image stands for generating foreground regions using the classification score maps. Cross-Image stands for our proposed CA-Classifier which leverages cross-images semantics.}
    \label{tab:design}
\end{table}

\begin{table}[t]
\centering
\small
\setlength{\tabcolsep}{5pt}
\footnotesize
{
    \begin{tabular}{cccc|ccccc}
        \toprule
        Size & All & Base & Novel & Top-K & All & Base & Novel\\
        \midrule
        \midrule
        10 & \textbf{32.6} & \textbf{42.8} & 26.2 & 10 & 31.6 & 40.3 & 26.2\\
        20 & 32.0 & 40.5 & \textbf{26.7} & 20 & \textbf{32.4} & 41.3 & \textbf{26.8}\\
        40 & 32.2 & 42.0 & 26.1 & 40 & 32.2 & \textbf{42.0} & 26.1\\
        80 & 31.6 & 40.9 & 25.7 & 80 & 31.4 & 40.0 & 25.9\\
        \bottomrule
    \end{tabular}
    }
    \caption{\small{Ablation on memory bank sizes and top-K region areas. We select the hyper-parameters with the best \texttt{novel} categories performance.}}
    \label{tab:memory_bank}
\end{table}

\noindent\textbf{Effectiveness of CA-Classifier.} We ablate the effectiveness of the CA-Classifier in the single- and multi-label classification data setting. As shown in Table~\ref{tab:ca_cls_effective}, the CA-Classifier improves the performance when applied to each type of classification data. The best performance is achieved when using a CA-Classifier on both types of classification data. 

\noindent\textbf{Design of CA-Classifier.} Another possible design is to simply select the high-score pixels in the classification score map as foreground regions. Specifically, we dynamically select foreground regions for a category and apply cross-entropy loss to the foreground regions. As shown in Table~\ref{tab:design}, our CA-Classifier outperforms the single-image design by $2.0$ mIoU on \texttt{novel} categories. The reason is that $\Omega_{c}$ sometimes incorrectly locates the frequently co-occurred background categories. 

\noindent\textbf{Ablation on Memory Bank.} We ablate different memory sizes $M$ with the fixed Top-K region area 40 in the left part of Table~\ref{tab:memory_bank}. LarvSeg has the highest mIoU on \texttt{novel} categories when $M=20$. Then, we ablate different top-$K$ with the fixed memory bank size 20 in the right part of Table~\ref{tab:memory_bank}. We select $K=20$ for the best \texttt{novel} category performance. 

\subsection{Qualitative Results}
\label{subsec:qualitative_results}

\begin{figure}[t]
    \centering
    \includegraphics[width=\linewidth]{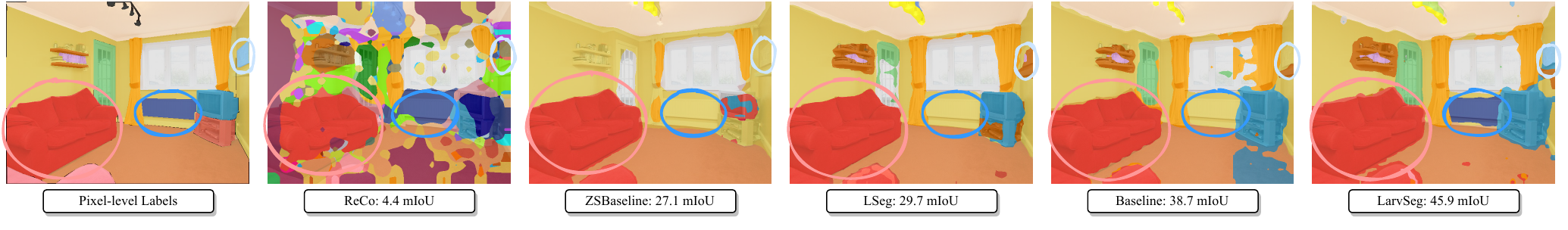}
    \caption{Visualization of model predictions. The tags show model names and the corresponding mIoUs of this image. Circles with different colours represent regions with \texttt{novel} categories in the image: \emph{sofa} (in the red circle), \emph{radiator} (in the dark blue circle) and \emph{painting} (in the light blue circle).
    }
    \label{fig:visualization}
\end{figure}

\begin{figure}[t]
    \centering
    \includegraphics[width=0.8\linewidth]{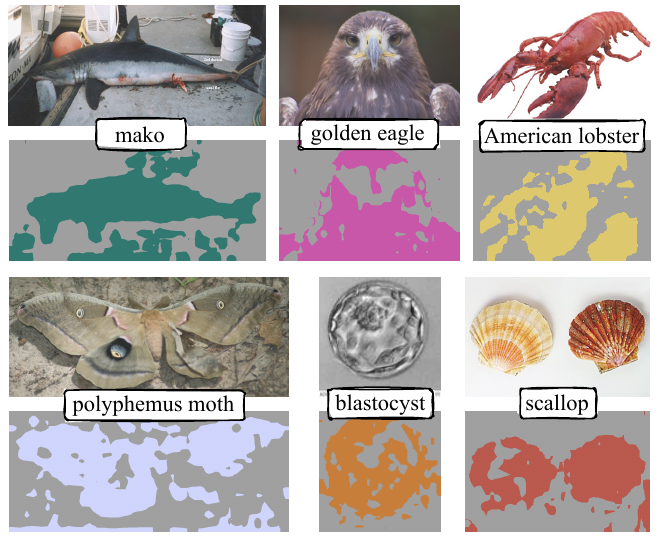}
    \caption{Visualization of 21K categories semantic segmentation.}
    \label{fig:in21k}
\end{figure}

\noindent\textbf{Segmentation on A150.} In Figure~\ref{fig:visualization}, we visualize the predictions of different models and highlight the appeared \texttt{novel} categories with circles. 
The \emph{sofa} is recognized by all models, the \emph{radiator} is recognized by ReCo and LarvSeg, and the \emph{painting} is only recognized by LarvSeg. 

\noindent\textbf{Segmentation For 21K Categories.} In Figure~\ref{fig:in21k}, we provide qualitative results of LarvSeg trained with C171 and I21K to show that our framework can extend semantic segmentation to 21K categories. We only visualize the predicted foreground masks and set all background predictions to grey.  Figure~\ref{fig:in21k} shows that the proposed LarvSeg is capable of segmenting fine-grained categories like \emph{polyphemus moth}, which demonstrates the superiority of the proposed framework.
More visualizations and details are provided in Supplementary Material.

\section{Conclusion}
\label{sec:conclusion}
In this paper, we address large vocabulary semantic segmentation using image classification data. We design a framework called \textbf{LarvSeg} to bridge the gap between image-level and pixel-level labels effectively. Firstly, We construct a simple baseline to incorporate image-level supervision which performs better than state-of-the-art language-guided semantic segmentation models. Then, we observe that a model trained on segmentation data can group the pixels of unseen categories as well. Based on this observation, we propose a category-wise attentive classifier to apply image-level supervision on the corresponding regions. Extensive experiments show that the proposed LarvSeg framework significantly outperforms the baseline model. For the first time, we provide a semantic segmentation model that can recognize 21K categories. We hope this new paradigm and the LarvSeg framework could be a strong baseline for large vocabulary semantic segmentation and facilitate future research.

\newpage
\bibliographystyle{splncs04}
\bibliography{reference} 

\end{document}